\pdfoutput=1
\PassOptionsToPackage{table}{xcolor}
\documentclass[sigconf,nonacm,acmthm=false]{acmart}

\usepackage{graphicx}
\usepackage{adjustbox}
\usepackage{amsmath}
\usepackage{float}
\makeatletter
\theoremstyle{plain}\newtheorem{theorem}{Theorem}[section]
\theoremstyle{definition}\newtheorem{definition}{Definition}[section]
\makeatother
\usepackage{booktabs}
\usepackage[table]{xcolor}
\usepackage{xspace}
\usepackage{hyperref}
\usepackage[normalem]{ulem}
\usepackage{url}
\usepackage{seqsplit}
\usepackage[T1]{fontenc}
\usepackage{cleveref}
\usepackage{tikz}
\usetikzlibrary{arrows.meta, positioning, shapes.geometric, decorations.pathreplacing, calc}
\usepackage{pgfplots}
\pgfplotsset{compat=1.18}
\usepgfplotslibrary{fillbetween}
\usepgfplotslibrary{groupplots}
\NewDocumentCommand{\replaced}{o+m+m}{#2}
\NewDocumentCommand{\deleted}{o+m}{}

\newcommand{\pth}{\,\text{--}\,}

\DeclareRobustCommand{\hfull}{\begin{tikzpicture}[baseline=-0.55ex]\fill (0,0) circle (0.52ex);\end{tikzpicture}}
\DeclareRobustCommand{\hhalf}{\begin{tikzpicture}[baseline=-0.55ex]\draw (0,0) circle (0.52ex);
  \fill (0,0) -- (90:0.52ex) arc (90:270:0.52ex) -- cycle;\end{tikzpicture}}
\DeclareRobustCommand{\hnone}{\begin{tikzpicture}[baseline=-0.55ex]\draw (0,0) circle (0.52ex);\end{tikzpicture}}

\newcommand{\sysname}{Anduin\xspace}

\NewDocumentCommand{\added}{o+m}{\ifpublish#2\else{\color{blue!65!black}#2}\fi}

\DeclareRobustCommand{\Rflow}{\ensuremath{\Pi_{\mathtt{flow}}}}
\DeclareRobustCommand{\Rsum}{\ensuremath{\Pi_{\mathtt{sum}}}}
\DeclareRobustCommand{\Ramp}{\ensuremath{\Pi_{\mathtt{amp}}}}

\DeclareRobustCommand{\Rcf}{\ensuremath{\Pi_{\mathtt{cf}}}}
\DeclareRobustCommand{\Ruse}{\ensuremath{\Pi_{\mathtt{use}}}}
\DeclareRobustCommand{\Rspend}{\ensuremath{\Pi_{\mathtt{spend}}}}

\newif\ifpublish
\publishtrue

\begin{document}

\title{\sysname: A Sybil-Resistant Agentic Marketplace Reputation System}

\author{Reza Ramji}
\authornote{This work was done while the author was at Circle.}
\affiliation{\institution{Circle Research}\country{}}
\author{Preston Vander Vos}
\affiliation{\institution{Circle Research}\country{}}
\author{Ziming Zeng}
\affiliation{\institution{Circle Research}\country{}}
\author{Duc V. Le}
\affiliation{\institution{Circle Research}\country{}}
\renewcommand{\shortauthors}{Ramji et al.}

\begin{abstract}
    As autonomous AI agents enter permissionless onchain markets, selecting new service providers remains a core challenge. Existing reputation systems depend on scarce resources like human curation, identity credentials, or voluntary reviews, while naive volume rankings are vulnerable to Sybil attacks.

We evaluate whether existing onchain payment histories can yield provider
ratings that are resilient against Sybil Attacks. Using a
$157$-day dataset comprising $66$M agentic transactions worth $\$6.1$M across 16
blockchains, we model payment flows as a bipartite graph between buyers and
providers. A provider’s rating is derived from graph paths linked to a buyer’s
existing trusted network. We propose three Sybil resistant rating algorithms: a
max-flow rating and two relaxed algorithms {optimized for stronger rating
performance}.

Evaluating these algorithms against simulated adversaries across $813$ buyers,
we show that rating manipulation costs $20$\% of those buyers' total spend
and is around {$23\times$} harder to manipulate than the standard similarity
rule (collaborative filtering).

 \end{abstract}

\maketitle

\section{Introduction}
\label{sec:introduction}

AI agents are transacting autonomously, selecting and paying for online services
to fulfill delegated tasks~\cite{coinbase2025x402,tempo2026machine}. 
Small, automated microtransactions suit traditional checkout rails
poorly~\cite{zhang2026sok,ling2026agentic}, making onchain stablecoins the
default settlement medium.

However, permissionless networks create a discovery dilemma: anyone can list an
endpoint with arbitrary descriptions, leaving buyers unable to distinguish
reliable providers from fraudulent or low-quality services. While human commerce
relies on brand reputation, reviews, and legal
recourse~\cite{resnick2002trust,nosko2015limits}, autonomous agents
cannot efficiently access these datapoints.
Current reputation solutions impose restrictive assumptions that undermine open
markets: curation of the marketplace {prohibits open participation} ~\cite{coinbase2026bazaar}, identity credentials
attest to entity identity rather than service
quality~\cite{choudhuri2026cryptographic}, escrowed evaluators can incur fees that
dwarf the sub-cent payments evaluated~\cite{crapis2026ercb}, and voluntary onchain ratings
are sparsely reported ($<4{,}000$ ratings
total)~\cite{xiong2026trustless}.

A naive alternative ranks providers by raw onchain usage—counting
transactions, counterparties, or volume~\cite{coinbase2026bazaar}. However,
usage metrics are not effective against Sybil attacks~\cite{douceur2002sybil}:
an adversary can fund its own wallets and loop self-payments to inflate its rank.
In active marketplaces, manipulating median reputation costs
just $\$0.0027$ ($1/259\text{th}$ of median volume)~\cite{xiong2026trustless}.
Heavyweight solutions impose costs incompatible with micropayments, while lightweight defaults offer no Sybil protection.

In this work, we address the research question: \emph{Can a reputation system
for agentic commerce be constructed purely from payments already settled onchain
{while rendering Sybil manipulation provably costly?}}

Defeating Sybil attacks is straightforward if protocols can be altered:
deposits, proof-of-work, or credentials make synthetic identities costly.
However, each requires participants to lock capital, burn compute, or obtain
attestations and {that both parties agree to a convention \emph{prior} to transacting}. We therefore {consider whether Sybil security guarantees can be attained}
strictly from existing payment records without protocol changes.

We show that a graph-based flow measurement indeed achieves this. We assemble a
continuous $157$-day settlement dataset from x402, the largest agentic
marketplace.
It comprises $66{,}073{,}201$ payments across $16$ blockchains between March
23 and August 26, 2026 from $162{,}882$ buyers and $50{,}390$ providers, totalling
$\$6.1\text{M}$ in volume at a median payment of $\$0.01$.

From the data, we construct a bipartite buyer-provider graph. Drawing on
graph-based reputation~\cite{page1998pagerank} and localized graph
defenses~\cite{alvisi2013sok,liao2025transaction}, our solution \sysname rates candidate
providers by routing network flow from providers the buyer already pays
(``known'' providers). Because every path originates from the buyer's
established payments, an adversary cannot fabricate reputation via synthetic
nodes; it must route capital into the buyer's trusted neighborhood,
economically bounding manipulation.

We present \sysname, a Sybil-resistant rating system for agentic commerce. Its construction and evaluation constitute five contributions:
\begin{enumerate}\setlength{\itemsep}{1.5pt}\setlength{\parskip}{0pt}
    \item \emph{Empirical Dataset.} A continuous $157$-day x402 settlement
      dataset across $16$ blockchains ($66.1\text{M}$ payments,
      $\$6.1\text{M}$ volume). 
    \item \emph{Graph Representation.} A bipartite graph model of settled
      payments that derives personalized trust directly from transaction
      topology without requiring compelled ratings or protocol changes.
    \item \emph{System Framework.} Criteria evaluating schemes jointly by
      predictive accuracy for future provider selection and bounded economic
      attack cost.
    \item \emph{Flow-Based Ratings.} Three graph flow algorithms ($\Rflow$, $\Rsum$,
      and $\Ramp$) with formal proofs establishing explicit expenditure lower bounds for Sybil attackers.
    \item \emph{Trace-Driven Evaluation.} Empirical evaluation showing that
      $\Ramp$ predicts a buyer's next provider first for $23.6\%$ of buyers (vs.
      $3.7\%$ for naive usage) while forcing a Sybil adversary to spend $20\%$
      of target volume (vs. $0.9\%$ for collaborative filtering).
\end{enumerate}

The rest of the paper is organized as follows: \Cref{sec:related-works} surveys related work, \Cref{sec:background} provides background on agentic payments, \Cref{sec:system-model} formalizes rating schemes, \Cref{sec:constructions} presents our constructions, \Cref{sec:methodology} evaluates them empirically, and \Cref{app:discussion} discusses implications.
 \section{Related Work}
\label{sec:related-works}

\textbf{Agentic marketplaces} are forums for autonomous buyers to engage
unfamiliar providers. Navigating this environment requires evaluating service
quality; however, every existing reputation mechanism imposes requirements that
permissionless markets lack. We survey existing mechanisms by their properties, {and the older family of graph-based reputation schemes which inform our design.}

\paragraph{Curation.}
A curated registry of vetted providers (e.g., x402
Bazaar~\cite{coinbase2026bazaar,circle2026agentstack}) offers Sybil resistance
but scales poorly as service volume outpaces human auditability, creating
dependence on a single authority's judgment and capacity to approve new providers.

\paragraph{Identity Credentials.}
Proof-of-personhood primitives~\cite{choudhuri2026cryptographic,lin2025baid}
bound Sybils by certifying identity, rather than service quality.
Moreover, human identity standards translate poorly to autonomous
software agents~\cite{otsuka2026identity,worldcoin2023worldid}.

\paragraph{Per-Transaction Evaluation.}
Third-party evaluator escrows under EIP-8183~\cite{crapis2026ercb} provide
quality checks but become prohibitively expensive for microtransactions, where
evaluator fees and minimum thresholds (e.g., $\$1.00$) dwarf typical $\$0.01$
payments~\cite{crapis2026ercdiscussion}.

\paragraph{Optional Ratings.}
Registry standards like EIP-8004~\cite{rossi2025ercb} allow optional post-transaction
ratings. However, active reporting is rare ($<4{,}000$
ratings)~\cite{xiong2026trustless}, and optional feedback remains
vulnerable to self-issued reviews.

\paragraph{Usage Ranking.}
Zero-overhead implemented approaches rank providers by transaction count, unique
counterparties, or aggregate volume~\cite{coinbase2026bazaar}. However, they
offer no Sybil defense~\cite{douceur2002sybil}: an adversary can loop funds
internally to inflate rank at fee cost
(\Cref{sec:strawmen}), manipulating median
scores for as little as $\$0.0027$~\cite{xiong2026trustless}.

\paragraph{Graph-Based Reputation and Max-Flow.}
{
Bazaar~\cite{post2011bazaar} proposed using max flow on traditional online markets to bound Sybil attacks. This mechanism classified a \emph{single} proposed
transaction as risky or safe by testing whether max-flow (across payment history) between buyer and seller exceeded the proposed transaction size. We want to discern between many new candidate providers, so flow \emph{values} must discriminate among them rather than generate a binary safety classification. Unbounded flow allows many candidates to achieve the same
maximum score, limiting the applicability of the mechanism as a rating. We therefore combine this max-flow approach with the literature consensus that localization can limit Sybils~\cite{cheng2005sybilproof,cao2012sybilrank,page1998pagerank}.
}

\paragraph{Summary Comparison.}
\Cref{tab:related} evaluates existing reputation paradigms across the central requirements of permissionless agentic commerce. Existing Sybil-resistant designs rely on external infrastructure, while easily implementable approaches lack Sybil protection. \sysname\ bridges this gap by operating strictly over settled payment history while enforcing {financial costs to manipulation}.

\begin{table*}[!tb]
\centering\small
\setlength{\tabcolsep}{4pt}
\begin{tabular}{lcccc}
\toprule
 & No New & Quality & Sybil- & Scales Without \\
Approach & Infrastructure & Signal & Resistant & Human Effort \\
\midrule
Curation~\cite{coinbase2026bazaar,circle2026agentstack}
  & \hnone\ (reviewers)    & \hfull      & \hfull & \hnone \\
Credentials~\cite{choudhuri2026cryptographic,lin2025baid}
  & \hnone\ (issuer)       & \hnone      & \hfull & \hfull \\
Evaluation~\cite{crapis2026ercb}
  & \hhalf\ (evaluators)   & \hfull      & \hfull & \hhalf\ (fee $>$ payment) \\
Optional Ratings~\cite{rossi2025ercb}
  & \hfull                 & \hhalf\ (sparse) & \hnone & \hfull \\
Usage Ranking~\cite{coinbase2026bazaar}
  & \hfull                 & \hhalf\ (popularity) & \hnone & \hfull \\
\midrule
\textbf{\sysname}
  & \hfull                 & \hfull      & \hfull & \hfull \\
\bottomrule
\end{tabular}
\caption{Comparison of reputation approaches for agentic marketplaces. \hfull\ offered, \hhalf\ partially offered, \hnone\ not offered.}
\label{tab:related}
\end{table*}
 \section{Background}
\label{sec:background}

We formalize the notation and graph model used throughout the paper.

\subsection{x402 Payment}
x402 embeds payments within HTTP exchanges~\cite{coinbase2025x402}: a server returns HTTP $402$~\emph{Payment Required} with pricing, and the buyer retries with a signed authorization header submitted onchain by a facilitator. Because settlement requires only a cryptographic signature rather than an interactive checkout, purchases occur autonomously without human intervention. Settlements occur in stablecoins (predominantly USDC) across EVM networks and Solana via designated facilitators or proxy contracts (for more details see \Cref{app:x402-identification}).
x402 carries the vast majority of machine payment volume ($\$6.1\text{M}$ vs $\approx\$270\text{K}$ for the second largest MPP~\cite{tempo2026machine,x402scan2026,mppscan2026machine}). Hence, we choose to evaluate our system on x402 data. 

\subsection{The Payment Graph}
\label{subsec:payment-graph}

Let $B=\{b_1,b_2,\dots\}$ denote \emph{buyers} and $P=\{p_1,p_2,\dots\}$ denote \emph{providers}.
The sets are disjoint, with any entity acting as both being modeled as two nodes.
Settled payments thus define a weighted bipartite graph
\[
    G \;=\; (B \cup P,\; E,\; w), \qquad E \subseteq B \times P,
\]
where an edge exists if $b$ paid $p$ within look-back window $\tau$. Because transactions flow strictly from $B$ to $P$, the bipartite structure preserves payment direction, and we subsequently treat $E$ as an undirected edge set with edge weights $w(b,p)$ reflecting volume over $\tau$.

\paragraph{Neighborhood of a query.}
A rating is computed relative to querying buyer $b \in B$. We define three derived sets:
\emph{known providers} $K(b) = \{\, p \in P \mid (b,p) \in E \,\}$ directly paid by $b$;
\emph{co-users} $C(b) = \{\, b' \in B\setminus\{b\} \mid \exists\, p \in K(b), (b',p) \in E \,\}$ sharing at least one known provider; and
\emph{candidate providers} $Q(b) = \{\, p' \in P\setminus K(b) \mid \exists\, b' \in C(b), (b',p') \in E \,\}$ reachable via co-users that $b$ has not previously paid ($Q(b) \cap K(b) = \emptyset$).

\begin{definition}[Route]
\label{def:route}
A \emph{route} from buyer $b$ to a candidate provider $p' \in Q(b)$ is a 3-hop path $b\pth p\pth b'\pth p'$ such that $p \in K(b)$ and $b' \in C(b)$. Its associated \emph{weight triple} is $\bigl(w(b,p),\,w(b',p),\,w(b',p')\bigr)$.
\end{definition}

{We confine our inference to this neighborhood; a
candidate provider is reached exclusively via a co-user.
Localizing trust to paths anchored at the buyer's own payments allows us to 
bound an adversary's influence.
We do this by picking routes from $K(b)$, so $b$ may infer trust to a candidate provider if and only if it buys from the original buyer's own known-provider.
We use this payment to a known provider to bound our budget according to \Cref{thm:budget}.%
\Cref{fig:notation} illustrates the complete structural model.

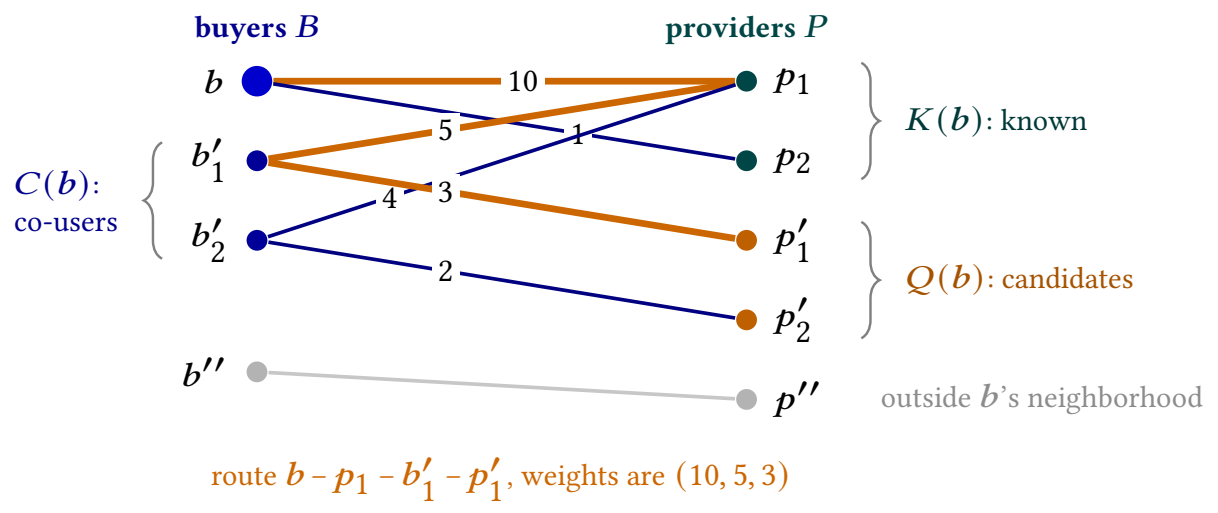
\begin{figure}[!tb]
\centering
\scalebox{1.15}{%
\begin{tikzpicture}[
  x=0.88cm, y=0.88cm,
  buyer/.style={circle, fill=blue!60!black, inner sep=0pt, minimum size=3.4pt},
  query/.style={circle, fill=blue!80!black, inner sep=0pt, minimum size=5pt},
  prov/.style={circle, fill=teal!55!black,  inner sep=0pt, minimum size=3.4pt},
  cand/.style={circle, fill=orange!75!black, inner sep=0pt, minimum size=3.4pt},
  other/.style={circle, fill=black!30, inner sep=0pt, minimum size=3.4pt},
  pay/.style={draw=blue!50!black, semithick},
  dim/.style={draw=black!22, semithick},
  rt/.style={draw=orange!80!black, line width=1.1pt},
  wt/.style={font=\tiny, inner sep=1pt, fill=white},
  nm/.style={font=\tiny, inner sep=1.5pt},
  hd/.style={font=\tiny\bfseries},
  br/.style={decorate, decoration={brace, amplitude=3pt}, draw=black!50},
  lb/.style={font=\tiny, align=left},
]
  \node[hd, blue!55!black] at (0,1.62)   {buyers $B$};
  \node[hd, teal!55!black] at (3.2,1.62) {providers $P$};

  \node[query] (b)  at (0,1.28)   {}; \node[nm, left=1pt of b]  {$b$};
  \node[buyer] (c1) at (0,0.76)   {}; \node[nm, left=1pt of c1] {$b'_1$};
  \node[buyer] (c2) at (0,0.24)   {}; \node[nm, left=1pt of c2] {$b'_2$};
  \node[other] (o)  at (0,-0.62)  {}; \node[nm, left=1pt of o]  {$b''$};

  \node[prov]  (p1) at (3.2,1.28) {}; \node[nm, right=1pt of p1] {$p_1$};
  \node[prov]  (p2) at (3.2,0.76) {}; \node[nm, right=1pt of p2] {$p_2$};
  \node[cand]  (q1) at (3.2,0.24) {}; \node[nm, right=1pt of q1] {$p'_1$};
  \node[cand]  (q2) at (3.2,-0.28){}; \node[nm, right=1pt of q2] {$p'_2$};
  \node[other] (p3) at (3.2,-0.80){}; \node[nm, right=1pt of p3] {$p''$};

  \draw[rt]  (b)  -- (p1) node[wt, pos=0.54] {$10$};
  \draw[pay] (b)  -- (p2) node[wt, pos=0.66] {$1$};
  \draw[rt]  (c1) -- (p1) node[wt, pos=0.38] {$5$};
  \draw[pay] (c2) -- (p1) node[wt, pos=0.26] {$4$};
  \draw[rt]  (c1) -- (q1) node[wt, pos=0.38] {$3$};
  \draw[pay] (c2) -- (q2) node[wt, pos=0.38] {$2$};
  \draw[dim] (o)  -- (p3);

  \draw[br] (3.95,1.40) -- (3.95,0.64) node[lb, midway, right=4pt, teal!45!black]
    {$K(b)$: known};
  \draw[br] (3.95,0.36) -- (3.95,-0.40) node[lb, midway, right=4pt, orange!65!black]
    {$Q(b)$: candidates};
  \draw[br] (-0.62,0.12) -- (-0.62,0.88) node[lb, midway, left=4pt, blue!55!black]
    {$C(b)$:\\co-users};

  \node[font=\tiny, black!45, anchor=west] at (3.95,-0.80) {outside $b$'s neighborhood};
  \node[font=\tiny, orange!80!black, anchor=north] at (1.6,-1.05)
    {route $b\pth p_1\pth b'_1\pth p'_1$, weights are $(10,5,3)$};
\end{tikzpicture}}
\caption{The bipartite payment graph model. For querying buyer $b$: $K(b)$ are known providers, $C(b)$ are co-users sharing known providers, and $Q(b)$ are candidate providers reachable through co-users. A 3-hop route $b\pth p\pth b'\pth p'$ carries three edge weights. Unconnected nodes ($b'', p''$) remain unrated.}
\label{fig:notation}
\end{figure}

 \section{Rating Schemes}
\label{sec:system-model}

This section defines a graph-based rating scheme, formalizes the adversary model, and establishes {our} evaluation criteria, \emph{utility} and \emph{attack cost}.

\subsection{Graph-Based Rating Schemes}
\label{sec:syntax}

A rating is a system by which a buyer may discern between similar candidate providers. It is a method for a buyer to leverage existing information (i.e. the payment graph) to decide which provider to purchase from next. Formally:

\begin{definition}[Rating Scheme]
\label{def:scheme}
A \emph{rating scheme} is an efficiently computable function:
\[
  \Pi \;:\; \mathcal{G} \times B \times P \;\longrightarrow\; \mathbb{R}_{\ge 0},
\]
where $\mathcal{G}$ is the set of weighted bipartite payment graphs $G=(B \cup P, E, w)$. Given graph $G$, querying buyer $b$, and candidate provider $p'$, the scheme outputs a non-negative score $\Pi_G(b,p')$. The recommendation list presented to $b$ consists of candidates $Q(b)$ ordered by decreasing score, with ties broken 
{at random}. 
We write $\mathrm{rank}_\Pi(b,p')$ for the 1-indexed position of $p'$ in this sorted list. 
\end{definition}

A rating scheme under \Cref{def:scheme} must identify high-quality providers (\emph{utility}) and make manipulation prohibitively expensive (\emph{attack cost}). Neither metric is meaningful in isolation: a constant rating is tamper-proof but uninformative, while {an accurate ranking that is trivially subverted is not useful in practice}.

\subsection{Threat Model}
\label{sec:threat}

An adversary is a malicious entity controlling addresses $A \subseteq B \cup P$ that seeks to inflate the rating of a target provider $p^{*} \in A$. Operating strictly over the payment graph $(E, w)$, the adversary's sole attack mechanism is edge creation under two assumptions:
\begin{itemize}
  \item[$A_1$:] \emph{Permissionless entry.} The adversary can create arbitrary buyer and provider identities (i.e.,  adding more elements to $A$) and transfer funds internally, each transaction costing base fee $\epsilon$.
  \item[$A_2$:] \emph{Uncompromised known providers.} For honest buyer $b$, the adversary controls no provider in $K(b)$.
\end{itemize}

\begin{definition}[Cost of Edge Creation]
\label{def:edge-cost}
For an adversary controlling address set $A$, creating edge $(b,p)$ with weight $w$ where $b \in A$ incurs cost:
\[
  \mathrm{cost}(b,p,w) \;=\;
  \begin{cases}
    \epsilon        & \text{if } p \in A,\\[2pt]
    w + \epsilon    & \text{if } p \notin A,
  \end{cases}
\]
where $\epsilon$ is the base onchain transaction and account generation fee.
\end{definition}

\noindent
As defined in \Cref{def:edge-cost} and illustrated in \Cref{fig:forgery}, value transferred between addresses within $A$ remains under adversarial control (except some fee $\epsilon$), whereas payments to external addresses are permanently forfeited. In microtransaction markets, base fees $\epsilon$ can be comparable to payment values, making security bounds functions of $\epsilon$. 

\begin{figure}[!tb]
\centering
\begin{tikzpicture}[
  x=1cm, y=1cm,
  q/.style={circle, fill=blue!85!black, inner sep=0pt, minimum size=5pt},
  b/.style={circle, fill=blue!60!black, inner sep=0pt, minimum size=3.8pt},
  p/.style={circle, fill=teal!55!black, inner sep=0pt, minimum size=3.8pt},
  s/.style={circle, fill=red!75!black, inner sep=0pt, minimum size=3.8pt},
  free/.style={draw=red!70, line width=0.9pt, dashed},
  paid/.style={draw=red!70, line width=0.9pt},
  no/.style={draw=black!35, line width=0.9pt, dotted},
  t/.style={font=\tiny, inner sep=2pt},
  lb/.style={font=\tiny, anchor=west, inner sep=3pt},
  grp/.style={decorate, decoration={brace, amplitude=5pt}, draw=black!60},
  glb/.style={font=\scriptsize, inner sep=4pt},
]
  \node[s] (a1) at (0,1.24) {};   \node[t, left=1pt of a1] {$b^{*}$};
  \node[s] (a2) at (1.2,1.24) {}; \node[t, right=1pt of a2] {$p^{*}$};
  \draw[free] (a1) -- (a2);
  \node[lb] at (1.9,1.24) {both ends adversarial: value returns, costs $\epsilon$};

  \node[s] (c1) at (0,0.62) {};   \node[t, left=1pt of c1] {$b^{*}$};
  \node[p] (c2) at (1.2,0.62) {}; \node[t, right=1pt of c2] {$p$};
  \draw[paid] (c1) -- (c2);
  \node[lb] at (1.9,0.62) {adversary pays an honest provider: costs the full weight $w+\epsilon$};

  \node[q] (d1) at (0,0.00) {};   \node[t, left=1pt of d1] {$b$};
  \node[p] (d2) at (1.2,0.00) {}; \node[t, right=1pt of d2] {$p$};
  \draw[no] (d1) -- (d2);
  \node[lb] at (1.9,0.00) {any edge with honest buyer $b$: \textbf{cannot be fabricated}};

  \draw[grp] (-0.95,0.34) -- (-0.95,1.52)
    node[glb, midway, left=5pt] {$A_1$};
  \draw[grp] (-0.95,-0.28) -- (-0.95,0.28)
    node[glb, midway, left=5pt] {$A_2$};
\end{tikzpicture}
\caption{Cost of forging an edge by type under $A_1$--$A_2$. Internal edges within $A$ cost only network fee $\epsilon$ while payments to external providers cost full value $w+\epsilon$ and edges incident to honest buyer $b$ cannot be fabricated at any price.}
\label{fig:forgery}
\end{figure}
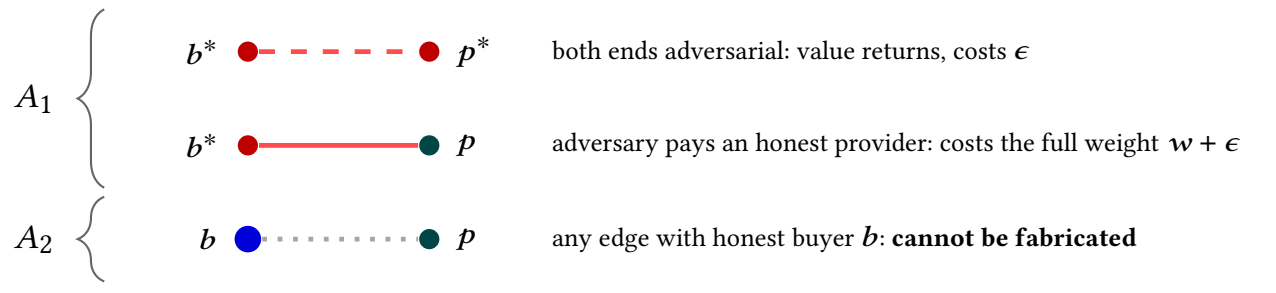
 
Finally, we write $G \rightsquigarrow G'$ when $G'$ is reachable from $G$ via adversarial edge and identity additions in $A$, and denote by $\chi(G,G')$ the total cumulative cost under \Cref{def:edge-cost}.

\subsection{Utility and Security Framework}
\label{sec:security}

\paragraph{Utility.}
Utility measures whether a scheme ranks the provider a buyer actually selects next higher than an unchosen alternative.
For a naive random ranking, the expected pairwise ranking accuracy is $\frac{1}{2}$, measuring global ranking across candidate providers $Q(b)$.
Our second metric, $P@k$, evaluates a stricter condition: whether the true provider falls within the top $k$ recommendations.
For evaluation, we fix a cutoff $T$ such that $G$ contains all payments settled strictly before $T$, and payments after $T$ are held out. The ground truth $p^+$ is the first provider outside $K(b)$ that buyer $b$ pays strictly after $T$.
\begin{definition}[Utility]
\label{def:utility}
Let $\mathcal{R} = \{(b_1, p^+_1), (b_2, p^+_2), \ldots\}$ denote the set of buyer--provider pairs where $b$ adopts a new provider $p^+ \notin K(b)$. For a pair $(b,p^+)$ sampled from $\mathcal{R}$ and a candidate provider $q$ drawn uniformly at random from $Q(b) \setminus \{p^+\}$, define utility:
\[
\begin{aligned}
    \mathrm{AUC} \;&:=\; {\Pr\bigl[\, \Pi_G(b,p^+) > \Pi_G(b,q) \,\bigr]
                +\tfrac{1}{2}\Pr\bigl[\, \Pi_G(b,p^+) = \Pi_G(b,q) \,\bigr]},\\[3pt]
    P@k \;&:=\; \Pr\bigl[\, \mathrm{rank}_\Pi(b,p^+) \le k \,\bigr].
\end{aligned}
\]
$\mathrm{AUC} = \frac{1}{2}$ or $P@k \approx 0$ indicate that a rating provides no predictive utility regarding new provider $p^+$.
\end{definition}

\paragraph{Attack Cost.}
Attack cost measures the cost for an adversary to elevate a target provider to a given position in a buyer's recommendation list. We formalize this as a minimum-cost optimization problem over adversary-constructable graphs.

\begin{definition}[Attack Cost]
\label{def:cost}
For an adversary provider $p^*$ and a querying buyer $b$, the adversary's minimum cost to achieve rank $k$ or higher is:
\[
    \min\;\bigl\{\, \chi(G,G') \;:\; G \rightsquigarrow G',\;\;
    \mathrm{rank}_{\Pi_{G'}}(b,p^{*}) \le k \,\bigr\}.
\]
\end{definition}
\noindent
Security against a Sybil attack in a permissionless marketplace takes the form of maximizing this attack cost.

 \section{Constructions}
\label{sec:constructions}

In this section, we describe \sysname's edge weighting, evaluate three naive baselines that fail under simple Sybil manipulation at cost $\epsilon$, and propose its three graph-based rating constructions.

\subsection{Choosing the Edge Weight}
\label{sec:weights}

\begin{definition}[Edge Weight Function]
\label{def:weight}
Let $T$ be a reference timestamp and $\mathcal{P}(b,p)=\{(t_1,v_1),(t_2,v_2),\dots\}$ denote the set of settled payments from buyer $b$ to provider $p$ at or before $T$, where each payment carries timestamp $t_i$ and value $v_i$ in USD.
Let $\delta:\mathbb{R}^{+}\rightarrow(0,1]$ be a monotonically decreasing time-decay function.
A \emph{weight function} maps $\mathcal{P}(b,p)$ to a positive number $w(b,p) \in \mathbb{R}^+$.
\end{definition}

\noindent
Because the transaction record lacks explicit feedback ratings, buyer preference must be inferred behaviorally~\cite{hu2008collaborative} via recency, frequency, and monetary volume~\cite{fader2005rfm}. We incorporate recency with decay $\delta$, which we instantiate as a step function over the window $\tau$ ($\delta(T-t_i)=1$ if $T-t_i\le\tau$, and $0$ otherwise), leaving value and frequency as the primary weighting factors. We instantiate four weighting function variants alongside an operational variant inclusive of gas fees:

\paragraph{Monetary Weight ($w_{\$}$).}
$w_{\$}(b,p) = \sum_i \delta(T-t_i)\,v_i$ evaluates decayed dollar volume. Inflating this weight requires {expending} capital at an uncompromised provider, establishing a direct financial attack cost.

\paragraph{Monetary Weight with Fees ($w_{\$\epsilon}$).}
$w_{\$\epsilon}(b,p) = \sum_i \delta(T-t_i)\,(v_i + \epsilon)$ incorporates onchain settlement fee $\epsilon$. Because $\epsilon$ is comparable to median microtransactions, $w_{\$\epsilon}$ acts as a volume-frequency hybrid. All empirical attack costs in \Cref{sec:methodology} are denominated in $w_{\$\epsilon}$ to include this cost.

\paragraph{Damped Monetary Weight ($w_{\log}$).}
$w_{\log}(b,p) = \log(1 + \sum_i v_i)$ dampens heavy-tailed payment distributions, preventing a single high-value transaction from dominating payment graph edges.

\paragraph{Transaction Frequency ($w_{\#}$).}
$w_{\#}(b,p) = \sum_i \delta(T-t_i)$ counts payment occurrences {irrespective} of monetary magnitude. {Notably,} manipulating $w_{\#}$ costs only transaction fee $\epsilon$.

\subsection{Baseline methods}
\label{sec:strawmen}

We examine three intuitive baselines to motivate \sysname's design. Each {is a rating scheme under} \Cref{def:scheme}, but can be subverted at a small cost.

\begin{figure*}[!tb]
\centering
\scalebox{1.15}{%
\begin{tikzpicture}[
  x=1cm, y=1cm,
  b/.style={circle, fill=blue!60!black, inner sep=0pt, minimum size=3pt},
  q/.style={circle, fill=blue!85!black, inner sep=0pt, minimum size=4.2pt},
  p/.style={circle, fill=teal!55!black, inner sep=0pt, minimum size=3pt},
  c/.style={circle, fill=orange!75!black, inner sep=0pt, minimum size=3.4pt},
  s/.style={circle, fill=red!75!black, inner sep=0pt, minimum size=3pt},
  e/.style={draw=blue!50!black, thin},
  a/.style={draw=red!70, thin},
  ai/.style={draw=red!70, thin, dashed},
  t/.style={font=\tiny, inner sep=1pt},
  ti/.style={font=\tiny\bfseries, inner sep=1pt},
  pr/.style={font=\tiny, red!70!black, inner sep=1pt},
]
\begin{scope}[shift={(0,0)}]
  \node[ti] at (0.75,1.30) {usage ranking};
  \node[c] (x) at (1.5,0.42) {}; \node[t, right=1pt of x] {$p'$};
  \foreach \y/\n in {0.72/a,0.42/b,0.12/c} \node[s] (\n) at (0,\y) {};
  \foreach \n in {a,b,c} \draw[ai] (\n) -- (x);
  \node[pr] at (0.75,-0.2) {any buyer counts};
  \node[pr] at (0.75,-0.46) {manipulation price: $n\epsilon$};
\end{scope}
\begin{scope}[shift={(3.05,0)}]
  \node[ti] at (0.75,1.30) {co-user count};
  \node[q] (u) at (0,0.86) {}; \node[t, left=1pt of u] {$b$};
  \node[p] (k) at (1.5,0.86) {}; \node[t, right=1pt of k] {$p$};
  \node[c] (x) at (1.5,0.06) {}; \node[t, right=1pt of x] {$p'$};
  \draw[e] (u) -- (k);
  \foreach \y/\n in {0.46/a,0.16/b} { \node[s] (\n) at (0,\y) {};
    \draw[a] (\n) -- (k); \draw[ai] (\n) -- (x); }
  \node[pr] at (0.75,-0.2) {small payments qualify};
  \node[pr] at (0.75,-0.46) {$2n\epsilon$};
\end{scope}
\begin{scope}[shift={(6.1,0)}]
  \node[ti] at (0.75,1.30) {co-user volume};
  \node[q] (u) at (0,0.86) {}; \node[t, left=1pt of u] {$b$};
  \node[p] (k) at (1.5,0.86) {}; \node[t, right=1pt of k] {$p$};
  \node[c] (x) at (1.5,0.06) {}; \node[t, right=1pt of x] {$p'$};
  \node[s] (a) at (0,0.36) {};
  \draw[e] (u) -- (k); \draw[a] (a) -- (k);
  \draw[ai, line width=1.2pt] (a) -- (x) node[pr, midway, above, sloped] {};
\node[pr] at (0.75,-0.2) {unbounded self payments};
  \node[pr] at (0.75,-0.46) {$2\epsilon$};
\end{scope}
\end{tikzpicture}}
\caption{The three counting rules, and what each costs an adversary. Following
\Cref{fig:forgery}: blue edges are honest payments, solid red edges are payments the adversary
makes to an honest provider (costing full value), and dashed red edges are self-dealing between
adversary-controlled addresses (costing only fees). All three rules fall for transaction fees
alone, because none charges for the fabricated edges: usage counts any payment, counting co-users
admits an identity on a payment of any size, and summing spend at $p'$ reads edges the adversary
pays to itself.}
\label{fig:ladder-steps}
\end{figure*}
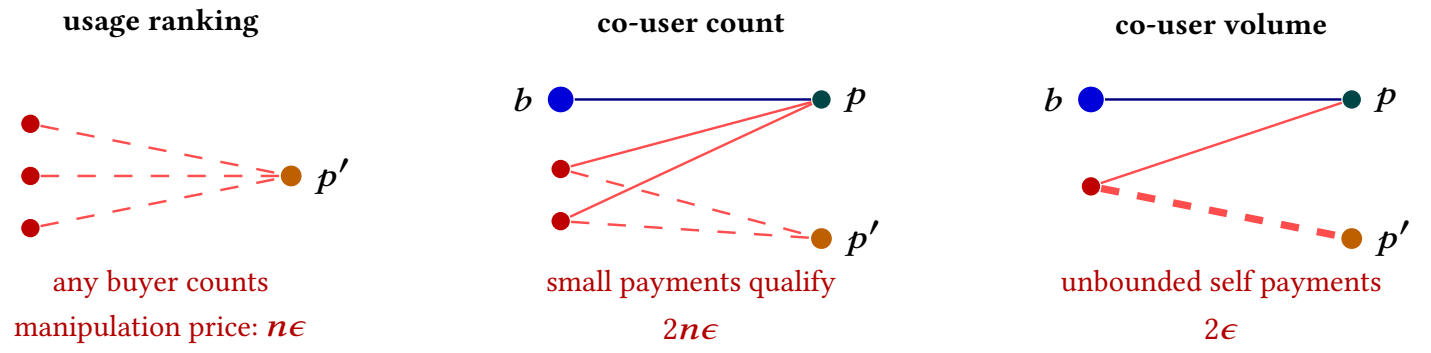
 
\paragraph{Usage Ranking ($\Ruse$).}
$\Ruse(b,p') = |\{\, b'' \in B \mid (b'',p') \in E \,\}|$ counts the number of distinct buyers who have paid $p'$. 
{Under $\Ruse$} an adversary owning $p'$ can execute $n$ self-directed payments at cost $n\epsilon$, inflating its rank without upper bound.

\paragraph{Co-User Count ($\Rcf$).}
$\Rcf(b,p') = |\{ b' \in C(b) \mid (b',p') \in E \}|$ implements collaborative filtering~\cite{linden2003amazon}, while $\Rcf$ counts co-user identities rather than payment volume. An adversary can generate $n$ Sybils, each making a micropayment to some $p \in K(b)$ to qualify as a co-user and another payment to $p'$ to register the edge, capturing top rank for cost $2n\epsilon$.

\paragraph{Co-User Volume ($\Rspend$).}
$\Rspend(b,p') = \sum_{b' \in C(b)} w(b',p')$ weights paths by volume into candidate provider $p'$. However, because $w(b',p')$ measures co-user payments into $p'$, an adversary {can execute a small payment to some $p\in K(b)$} and a large self-payment to $p'$, outranking honest candidates for cost $2\epsilon$.

\subsection{Maximum Flow Rating: $\Rflow$}
\label{sec:flow}

Baseline mechanisms fail because {they infer based on features of the graph which an} adversary can easily manipulate.
To guarantee that Sybils cannot inject unearned trust, {we constructed $\Rflow$ such that} \emph{no candidate provider can receive more trust than the {querying} buyer $b$  settled with its known providers}. This system maps reputation to a maximum-flow problem over a localized query network.

\begin{definition}[Query Network]
\label{def:network}
For querying buyer $b$ and candidate provider $p' \in Q(b)$, the \emph{query network} $N(b,p')$ is the directed flow network obtained by orienting every route $b\pth p\pth b'\pth p'$ as $b \to p \to b' \to p'$, with arc capacities $w(b,p)$, $w(b',p)$, and $w(b',p')$
{respectively.}
\end{definition}

\noindent
Our first score $\Rflow$ computes maximum network flow over $N(b,p')$:
\begin{equation}
  \fbox{$\displaystyle
    \Rflow(b,p') \;:=\; \operatorname{maxflow}\bigl(N(b,p'),\, b \to p'\bigr)
  $}
  \label{eq:rflow}
\end{equation}

\noindent
The capacity of {each} cut in $N(b,p')$ bounds the maximum flow from $b$ to $p'$.
By assumption $A_2$, the adversary cannot alter arcs incident to $b$, so the cut formed by arcs $\{b \to p \mid p \in K(b)\}$ is immutable.
Its total capacity $\beta(G,b) = \sum_{p \in K(b)} w(b,p)$ therefore provides an adversary-resistant upper bound on $\Rflow(b,p')$ (formally proven in \Cref{thm:spend}).
This rating, however, can limit differentiation.
If multiple candidate providers have maximum flow of $\beta(G,b)$, they score equally and cannot be ranked.

\subsection{Summation Flow Rating: $\Rsum$}
\label{sec:summation}

While $\Rflow$ enforces strict security bounds, co-users share the same source arcs out of $b$.
Therefore, routing flow through one reduces the capacity available to others.
In recommendation systems, however, multiple independent co-users choosing the same candidate {suggests additional} quality.
{$\Rsum$ implements this by allowing buyer capacity for each co-user.}

The \emph{bridge capacity} $\mathrm{bridge}(b,b') = \sum_{p \in K(b)} \min\bigl(w(b,p),\, w(b',p)\bigr)$ measures the aggregate two-hop capacity from $b$ to co-user $b'$ through their shared known providers.

Considering a single co-user $b'$, the paths through shared known providers are parallel and merge at $b'$. Therefore, total flow is limited by the bridge capacity on one side and the arc $b' \to p'$ on the other: $\min(\mathrm{bridge}(b,b'),\, w(b',p'))$.
In the shared network $N(b,p')$, co-users compete for the same source arcs out of $b$, so the combined flow is at most the sum of these per-co-user values.
$\Rsum$ realizes this sum exactly by treating each co-user independently, yielding $\Rsum(b,p') \ge \Rflow(b,p')$.
Summing across co-users gives the $\Rsum$ rating:

\begin{equation}
  \fbox{$\displaystyle
    \Rsum(b,p') \;:=\; \sum_{b' \,\in\, C(b)} \min\Bigl( \mathrm{bridge}(b,b'),\; w(b',p') \Bigr)
  $}
  \label{eq:rsum}
\end{equation}

\subsection{Amplified Flow Rating: $\Ramp$}
\label{sec:amplified}

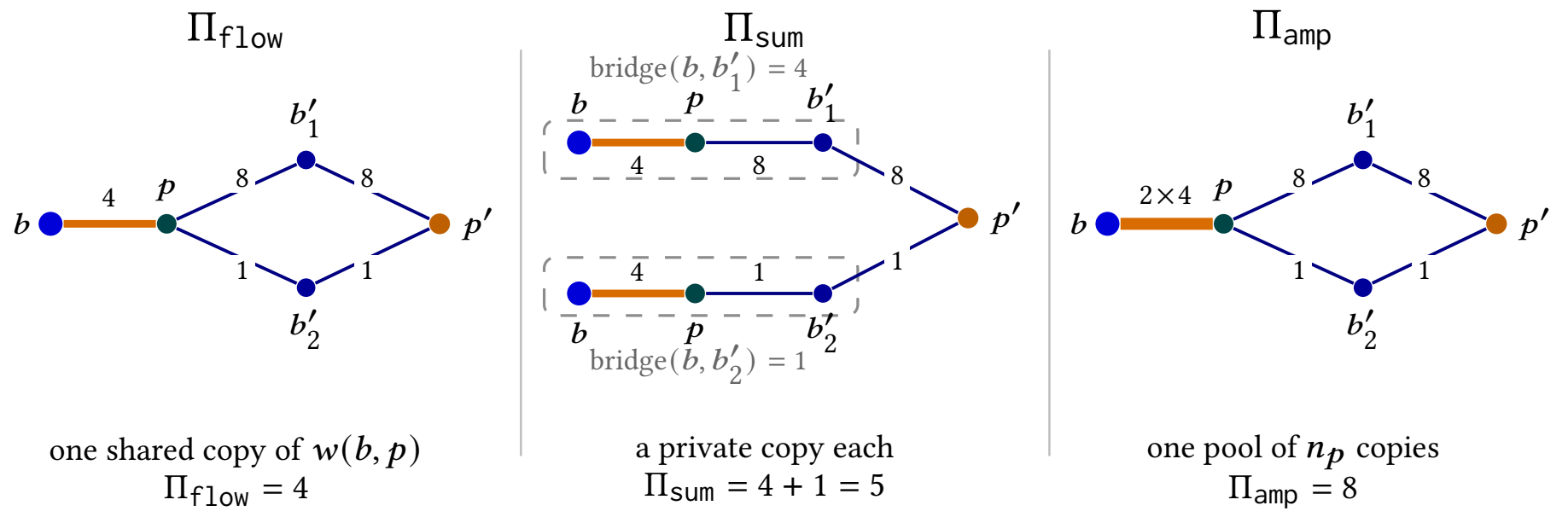
\begin{figure*}[!tb]
\centering
\scalebox{1.2}{\begin{tikzpicture}[
  x=0.75cm, y=0.75cm,
  q/.style={circle, fill=blue!85!black, inner sep=0pt, minimum size=4.4pt},
  b/.style={circle, fill=blue!60!black, inner sep=0pt, minimum size=3.4pt},
  p/.style={circle, fill=teal!55!black, inner sep=0pt, minimum size=3.6pt},
  c/.style={circle, fill=orange!75!black, inner sep=0pt, minimum size=3.8pt},
  e/.style={draw=blue!50!black, semithick},
  hi/.style={draw=orange!85!black, line width=1.4pt},
  wt/.style={font=\tiny, inner sep=1pt},
  ew/.style={font=\tiny, inner sep=1pt, fill=white},
  ti/.style={font=\small\bfseries},
  vl/.style={font=\scriptsize},
  bx/.style={draw=black!45, dashed, rounded corners=2pt, line width=0.5pt},
]
  \node[ti] at (1.6,2.72) {$\Rflow$};
  \node[q] (b)  at (0,1.05)    {}; \node[wt, left=0pt of b]  {$b$};
  \node[p] (p)  at (1.0,1.05)  {}; \node[wt, above=1pt of p] {$p$};
  \node[b] (c1) at (2.2,1.60)  {}; \node[wt, above=1pt of c1] {$b'_1$};
  \node[b] (c2) at (2.2,0.50)  {}; \node[wt, below=1pt of c2] {$b'_2$};
  \node[c] (x)  at (3.35,1.05) {}; \node[wt, right=1pt of x] {$p'$};
  \draw[hi] (b) -- (p);  \node[ew] at (0.5,1.28) {$4$};
  \draw[e] (p) -- (c1) node[ew, pos=0.6, above left=-2pt] {$8$};
  \draw[e] (p) -- (c2) node[ew, pos=0.6, below left=-2pt] {$1$};
  \draw[e] (c1) -- (x) node[ew, pos=0.4, above right=-2pt] {$8$};
  \draw[e] (c2) -- (x) node[ew, pos=0.4, below right=-2pt] {$1$};
  \node[vl, anchor=north, align=center] at (1.6,-0.62)
    {one shared copy of $w(b,p)$\\$\Rflow=4$};

  \draw[black!25] (4.05,-0.95) -- (4.05,2.55);

  \begin{scope}[shift={(4.55,0)}]
  \node[ti] at (1.6,2.72) {$\Rsum$};
  \node[q] (B1) at (0,1.75)   {};
  \node[p] (P1) at (1.0,1.75) {};
  \node[b] (D1) at (2.1,1.75) {};
  \draw[hi] (B1) -- (P1); \node[ew] at (0.50,1.56) {$4$};
  \draw[e]  (P1) -- (D1); \node[ew] at (1.55,1.56) {$8$};
  \draw[bx] (-0.30,1.44) rectangle (2.40,1.94);
  \node[wt] at (0,2.09) {$b$};  \node[wt] at (1.0,2.09) {$p$};
  \node[wt] at (2.1,2.09) {$b'_1$};
  \node[wt, black!60] at (1.03,2.36) {$\mathrm{bridge}(b,b'_1)=4$};
  \node[q] (B2) at (0,0.45)   {};
  \node[p] (P2) at (1.0,0.45) {};
  \node[b] (D2) at (2.1,0.45) {};
  \draw[hi] (B2) -- (P2); \node[ew] at (0.50,0.64) {$4$};
  \draw[e]  (P2) -- (D2); \node[ew] at (1.55,0.64) {$1$};
  \draw[bx] (-0.30,0.26) rectangle (2.40,0.76);
  \node[wt] at (0,0.11) {$b$};  \node[wt] at (1.0,0.11) {$p$};
  \node[wt] at (2.1,0.11) {$b'_2$};
  \node[wt, black!60] at (1.03,-0.16) {$\mathrm{bridge}(b,b'_2)=1$};
  \node[c] (X) at (3.35,1.10) {}; \node[wt, right=1pt of X] {$p'$};
  \draw[e] (D1) -- (X) node[ew, pos=0.5, above right=-3pt] {$8$};
  \draw[e] (D2) -- (X) node[ew, pos=0.5, below right=-3pt] {$1$};
  \node[vl, anchor=north, align=center] at (1.6,-0.62)
    {a private copy each\\$\Rsum=4+1=5$};
  \end{scope}

  \draw[black!25] (8.60,-0.95) -- (8.60,2.55);

  \begin{scope}[shift={(9.10,0)}]
  \node[ti] at (1.6,2.72) {$\Ramp$};
  \node[q] (G)  at (0,1.05)   {}; \node[wt, left=0pt of G] {$b$};
  \node[p] (Q)  at (1.0,1.05) {}; \node[wt, above=1pt of Q] {$p$};
  \node[b] (E1) at (2.2,1.60) {}; \node[wt, above=1pt of E1] {$b'_1$};
  \node[b] (E2) at (2.2,0.50) {}; \node[wt, below=1pt of E2] {$b'_2$};
  \node[c] (Y)  at (3.35,1.05) {}; \node[wt, right=1pt of Y] {$p'$};
  \draw[hi, line width=2.2pt] (G) -- (Q); \node[ew] at (0.5,1.30) {$2\!\times\!4$};
  \draw[e] (Q) -- (E1) node[ew, pos=0.6, above left=-2pt] {$8$};
  \draw[e] (Q) -- (E2) node[ew, pos=0.6, below left=-2pt] {$1$};
  \draw[e] (E1) -- (Y) node[ew, pos=0.4, above right=-2pt] {$8$};
  \draw[e] (E2) -- (Y) node[ew, pos=0.4, below right=-2pt] {$1$};
  \node[vl, anchor=north, align=center] at (1.6,-0.62)
    {one pool of $n_p$ copies\\$\Ramp=8$};
  \end{scope}
\end{tikzpicture}}
\caption{Rating comparison on an example ($w(b,p)=4$, $b'_1$ pays $p$ and $p'$ $\$8$ each, $b'_2$ pays $\$1$ each). $\Rflow$ shares $w(b,p)$, bottlenecking flow at $b$'s outgoing arc ($\Rflow=4$). $\Rsum$ allocates private copies per co-user, summing independent flows ($\Rsum=4+1=5$). $\Ramp$ pools $n_p=2$ copies across co-users paying $p'$, allowing $b'_1$ to draw all available capacity ($\Ramp=8$).}
\label{fig:three-ratings}
\end{figure*}
 
Under $\Rsum$, each co-user's contribution to $p'$ is capped at $w(b',p')$, so a co-user with high bridge capacity but small payments to $p'$ cannot fully use its capacity.
This underuse of information suggests a slightly suboptimal rating capability.
As illustrated in \Cref{fig:three-ratings}, $\Ramp$ pools this capacity across co-users who have transacted with candidate $p'$, allowing co-users with higher candidate provider payment volume to contribute more evidence.

Let $N_a(b,p')$ be a modified query network where the capacity of each arc $b \to p$ is scaled to $n_p \cdot w(b,p)$ where $n_p = |\{\, b' \in C(b) \mid (b',p) \in E \wedge (b',p') \in E \,\}|$. Define the amplified rating $\Ramp$:

\begin{equation}
  \fbox{$\displaystyle
    \Ramp(b,p') \;:=\; \operatorname{maxflow}\bigl(N_a(b,p'),\, b \to p'\bigr)
  $}
  \label{eq:ramp}
\end{equation}

\subsection{Computational Complexity}
\label{sec:complexity}

For $\Rsum$, $\mathrm{bridge}(b,b')$ depends only on $b'$ and not on the candidate provider, so it can be precomputed in $O(|K(b)|\cdot|C(b)|)$ time, yielding $O(1)$ evaluation per candidate.
In contrast, $\Rflow$ and $\Ramp$ run Dinic's max-flow algorithm~\cite{dinitz1970algorithm} once per candidate provider on the {\emph{localized} query network $N(b,p')$ (not the global graph) which has $V = |K(b)| + |C(b)| + 2$ nodes and $E = O(|K(b)|\cdot|C(b)|)$ arcs, so each candidate costs $O(V^2 E)$}.
\subsection{Security Guarantees}
\label{sec:why-replication}

We establish two core security bounds governing adversarial manipulation across our three constructions:

\begin{theorem}[Budget-Bounded Rating Cap]
\label{thm:budget}
Let $G \rightsquigarrow G'$ be a graph transformation where the adversary spends $m$ dollars in payments to external providers outside $A$. Under assumptions $A_1$ and $A_2$, $\Pi(b,p^*) \le m$ for $\Pi \in \{\Rflow, \Rsum, \Ramp\}$ regardless of the number of Sybil identities created. This bound is tight.
\end{theorem}
\begin{proof}
Let $A \subseteq B \cup P$ denote the set of adversary-controlled addresses, with target provider $p^* \in A$.
Under assumption $A_2$, the querying buyer's trusted provider set $K(b)$ satisfies $K(b) \cap A = \emptyset$.
Since $p^*$ is adversary-controlled and has not provided real service to honest buyers, no honest co-user $b' \notin A$ has paid $p^*$; any such payment would reflect genuine service, not manipulation.
Therefore, every route $b \to p \to b' \to p^*$ passes through $b' \in A \cap C(b)$ and only adversarial co-users contribute flow toward $p^*$.

Under \Cref{def:edge-cost}, each payment from an adversarial $b' \in A$ to an honest provider $p \notin A$ costs $w(b',p) + \epsilon$, permanently removing capital from $A$.
Let $m = \sum_{b' \in A} \sum_{p \in K(b)} w(b',p)$ denote the adversary's total payment volume directed to honest known providers.
Fees $\epsilon$ are excluded because they do not contribute to edge weight, so their inclusion would only increase true spend.

We consider each construction:
\begin{enumerate}
  \item $\Rflow$: In the query network $N(b,p^*)$, the boundary arcs $\{(p, b') \mid p \in K(b),\, b' \in A \cap C(b)\}$ form a cut separating $b$ from $p^*$ with total capacity:
  \[
    \sum_{b' \in A \cap C(b)} \sum_{p \in K(b)} w(b',p) \;=\; m.
  \]
  By the Max-Flow Min-Cut Theorem, maximum flow is bounded by the capacity of any cut; since this cut has capacity $m$, we have $\Rflow(b,p^*) \le m$.

  \item $\Rsum$: By definition, \[\Rsum(b,p^*) = \sum_{b' \in C(b)} \min(\mathrm{bridge}(b,b'), w(b',p^*)).\] Honest co-users $b' \notin A$ have never transacted with adversarial provider $p^*$, so $w(b',p^*) = 0$ for all $b' \notin A$. Thus, only adversarial identities $b' \in A$ contribute:
  \[
    \begin{aligned}\Rsum(b,p^*) &= \sum_{b' \in A \cap C(b)} \min\bigl(\mathrm{bridge}(b,b'),\, w(b',p^*)\bigr) \\ &\le \sum_{b' \in A \cap C(b)} \mathrm{bridge}(b,b').\end{aligned}
  \]
  Recall the bridge definition:
  \[
    \mathrm{bridge}(b,b') \;=\; \sum_{p \in K(b)} \min\bigl(w(b,p),\, w(b',p)\bigr) \;\le\; \sum_{p \in K(b)} w(b',p).
  \]
  Summing across all $b' \in A$:
  \[
    \Rsum(b,p^*) \;\le\; \sum_{b' \in A \cap C(b)} \sum_{p \in K(b)} w(b',p) \;=\; m.
  \]

  \item $\Ramp$: In $N_a(b,p^*)$, arc capacities leaving $b$ are scaled by $n_{p^*}$. However, the cut separating $K(b)$ from $A$ consists of the boundary arcs $(p, b')$ for $b' \in A$, whose capacities remain $w(b',p)$ unscaled. Because these arcs form a cut separating $b$ from $p^*$ (similar to $\Rflow$ case above), max-flow cannot exceed their aggregate capacity $m$. Hence $\Ramp(b,p^*) \le m$.
\end{enumerate}

To prove tightness, consider an adversary that instantiates $n$ Sybil buyers $\{b'_1, \dots, b'_n\} \subset A$, each paying an honest provider $p \in K(b)$ an amount $m/n$, and executing a self-directed payment $w(b'_i, p^*) \ge m/n$.
When $m/n \le w(b,p)$, $\min(w(b,p), w(b'_i,p)) = m/n$.
The bridge capacity for each identity is $\mathrm{bridge}(b,b'_i) = m/n$.
Summing across all $n$ identities yields $\sum_{i=1}^n (m/n) = m$, achieving the bound with equality.
\end{proof}

\begin{theorem}[Spend-Bounded Cap]
\label{thm:spend}
For any adversarially reachable graph $G \rightsquigarrow G'$, $\Rflow$ evaluated on $G'$ is bounded by buyer $b$'s total historical spend on $G$: $\Rflow(b,p') \le \beta(G,b) = \sum_{p \in K(b)} w(b,p)$ for all $p' \in Q(b)$.
\end{theorem}
\begin{proof}
The arcs $\{(b,p) \mid p \in K(b)\}$ form a cut in $N(b,p')$ separating $b$ from any candidate provider $p'$, with total capacity:
\[
  \sum_{p \in K(b)} w(b,p) \;=\; \beta(G,b).
\]
By the Max-Flow Min-Cut Theorem, maximum flow is bounded by the capacity of any cut, so $\Rflow(b,p') \le \beta(G,b)$.
Since settled transactions are immutably recorded on the public ledger, no adversary can alter or fabricate payments from $b$, so $\beta(G,b)$ is unchanged across all reachable graphs $G'$.
\end{proof}

These theorems establish two structural cap categories.
\Cref{thm:budget} gives an \emph{outlay-proportional cap} $m$ for all three schemes: reputation inflation requires proportional real expenditure, so Sybil identities paying only base fees $\epsilon$ provide zero rating leverage.
\Cref{thm:spend} gives a stricter \emph{buyer cap} $\beta(G,b)$ for $\Rflow$, determined solely by $b$'s historical spend and unimpacted by adversarial budget.

 \section{Evaluation}
\label{sec:methodology}

We evaluate \sysname on real-world transactions across two dimensions: ranking accuracy (\emph{utility}) and resilience against adversarial distortion (\emph{attack cost}).

\subsection{\sysname overview}
\label{sec:p-method}

\sysname takes the settlement ledger as input and produces a ranked candidate provider list for each querying buyer in four steps.

\paragraph{(1) Collecting Corpus.}
Following the criteria in \Cref{app:x402-identification}, we assemble onchain x402 stablecoin settlements (USDC/USDT) across 16 EVM networks and Solana over 157 days (March 23 to August 26, 2026), totaling $66.1\text{M}$ payments and $\$6.1\text{M}$.
Self-payments are removed.

\paragraph{(2) Snapshot and Population.}
We compute ratings over a 90-day look-back window ($\tau = 90$ days) ending June 21, 2026, 00:00 UTC.
This window includes $24{,}429{,}942$ distinct payments totaling $\$3.83\text{M}$.
We identify $136{,}646$ buyers and $3{,}649$ \emph{mature providers} (paid by $\ge 2$ buyers).
Our evaluation cohort comprises $813$ buyers who subsequently adopt a previously uncontacted mature provider.
{These are the only buyers with a ground-truth label: of the $136{,}646$, just $3{,}656$ transact again in the following 30 days, and $813$ of those first pay a mature provider they had never paid before.}

\paragraph{(3) Candidate Generation.}
Candidate providers $Q(b)$ comprise all mature providers reachable via at least one co-user that the buyer has never paid.
Across all buyers, lists average $381.4$ candidate providers, with a median of $370$ and a maximum of $747$.
Entering the top 25 requires outranking $93\%$ of candidate providers.

\paragraph{(4) Measurement Framework.}
Utility is evaluated by checking whether each buyer's actual next provider choice---the first new provider it pays in the 30 days after the cutoff---is ranked highly by the rating.
Attack cost (\Cref{sec:threat}) is measured by injecting a target provider $p^*$ into the snapshot; the attacker may then introduce arbitrarily many Sybil buyer identities that transact with the buyer's known providers, building the graph needed to elevate $p^*$'s ranking.
We solve for the minimum spend required for $p^*$ to achieve target rank $k$, { as illustrated in \Cref{fig:method}.}

\begin{figure*}[!tb]
\centering
\scalebox{1.2}{\begin{tikzpicture}[
  x=1cm, y=1cm,
  q/.style={circle, fill=blue!85!black, inner sep=0pt, minimum size=4.2pt},
  b/.style={circle, fill=blue!60!black, inner sep=0pt, minimum size=3.0pt},
  p/.style={circle, fill=teal!55!black, inner sep=0pt, minimum size=3.0pt},
  c/.style={circle, fill=orange!75!black, inner sep=0pt, minimum size=3.0pt},
  s/.style={circle, fill=red!75!black, inner sep=0pt, minimum size=3.0pt},
  tr/.style={circle, fill=green!45!black, inner sep=0pt, minimum size=4pt},
  e/.style={draw=blue!50!black, semithick},
  a/.style={draw=red!70, semithick},
  fut/.style={draw=green!45!black, semithick, dashed},
  t/.style={font=\tiny, inner sep=1.1pt},
  ti/.style={font=\scriptsize\bfseries},
  lab/.style={font=\tiny, text=black!55, inner sep=1pt},
  g/.style={draw=black!40, dashed, rounded corners=2pt},
  ar/.style={-{Latex[length=2mm]}, draw=black!55, thick},
]
  \draw[g] (-0.35,-0.75) rectangle (2.50,0.95);
  \node[ti] at (1.07,1.16) {1. Snapshot at $T$};
  \node[q] (u)  at (0.00,0.35) {}; \node[t, left=1pt of u] {$b$};
  \node[p] (p1) at (0.82,0.68) {}; \node[p] (p2) at (0.82,0.02) {};
  \node[b] (c1) at (1.58,0.68) {}; \node[b] (c2) at (1.58,0.02) {};
  \node[c] (x1) at (2.28,0.68) {}; \node[c] (x2) at (2.28,0.02) {};
  \draw[e] (u)--(p1); \draw[e] (u)--(p2); \draw[e] (c1)--(p1); \draw[e] (c2)--(p2);
  \draw[e] (c1)--(x1); \draw[e] (c2)--(x2);
  \node[t, black!60] at (1.07,-0.97) {$90$ days of real payments};

  \draw[ar] (2.62,0.46) -- (3.60,1.15) node[midway, above, sloped, lab] {utility};
  \draw[ar] (2.62,0.24) -- (3.60,-1.15) node[midway, below, sloped, lab] {attack};

  \draw[g] (3.65,0.55) rectangle (5.55,1.78);
  \node[ti] at (4.60,1.99) {2. Rank $Q(b)$};
  \foreach \y/\lbl in {1.55/1, 1.33/2, 1.11/3, 0.85/{$\vdots$}}
    { \fill[teal!55!black] (3.95,\y) circle (1.3pt); \node[t, right=2pt] at (4.05,\y) {\lbl}; }
  \node[t, black!60] at (4.60,0.38) {{hundreds of candidates}};

  \draw[ar] (5.72,1.16) -- (6.28,1.16);

  \draw[g] (6.40,0.55) rectangle (8.78,1.78);
  \node[ti] at (7.59,1.99) {3. What $b$ did after $T$};
  \node[q]  (v)  at (6.80,1.16) {}; \node[t, left=1pt of v] {$b$};
  \node[tr] (tt) at (8.28,1.16) {}; \node[t, right=1pt of tt, green!35!black] {$p^+$};
  \draw[fut] (v) -- (tt);
  \node[t, black!60] at (7.59,0.38) {was $p^+$ ranked first?};

  \draw[g] (3.65,-1.95) rectangle (6.7,-0.55);
  \node[ti] at (5.10,-0.34) {2. Inject adversary};
  \node[q] (w)  at (4.00,-1.02) {}; \node[t, left=1pt of w] {$b$};
  \node[p] (r1) at (4.82,-0.78) {}; \node[p] (r2) at (4.82,-1.34) {};
  \node[b] (d1) at (5.58,-0.78) {}; \node[b] (d2) at (5.58,-1.34) {};
  \node[c] (y1) at (6.28,-0.78) {};
  \draw[e] (w)--(r1); \draw[e] (w)--(r2); \draw[e] (d1)--(r1); \draw[e] (d2)--(r2);
  \draw[e] (d1)--(y1);
  \node[s] (s1) at (5.58,-1.60) {}; \node[s] (s2) at (5.58,-1.84) {};
  \node[s] (ps) at (6.28,-1.72) {}; \node[t, right=1pt of ps, red!70!black] {$p^{*}$};
  \draw[a] (s1)--(r2); \draw[a] (s2)--(r2); \draw[a] (s1)--(ps); \draw[a] (s2)--(ps);
  \node[t, red!70!black] at (5.10,-2.14) {Sybil identities, minimize spend};

  \draw[ar] (6.78,-1.05) -- (7.28,-1.05);

  \draw[g] (7.40,-1.95) rectangle (9.55,-0.55);
  \node[ti] at (8.47,-0.34) {3. Rescore, rank};
  \foreach \y/\col/\lbl in {-0.85/{teal!55!black}/1, -1.12/{teal!55!black}/2,
                            -1.39/{red!75!black}/3, -1.66/{teal!55!black}/4}
    { \fill[\col] (7.70,\y) circle (1.3pt); \node[t, right=2pt] at (7.80,\y) {rank \lbl}; }
  \node[t, black!60] at (8.47,-2.14) {does $p^{*}$ reach top $25$?};
\end{tikzpicture}}
\caption{Evaluation pipeline. From one snapshot of payments settled before $T$ we measure
both properties. \emph{Utility} (top): the scheme ranks the candidates $Q(b)$, and we check where
it places $p^+$, the first new provider $b$ actually pays after $T$. \emph{Attack cost} (bottom): we inject target provider $p^{*}$ and arbitrarily many Sybil identities that pay into $b$'s known providers, rescore, and record the minimum spend that lifts $p^{*}$ to rank $k$.}
\label{fig:method}
\end{figure*}
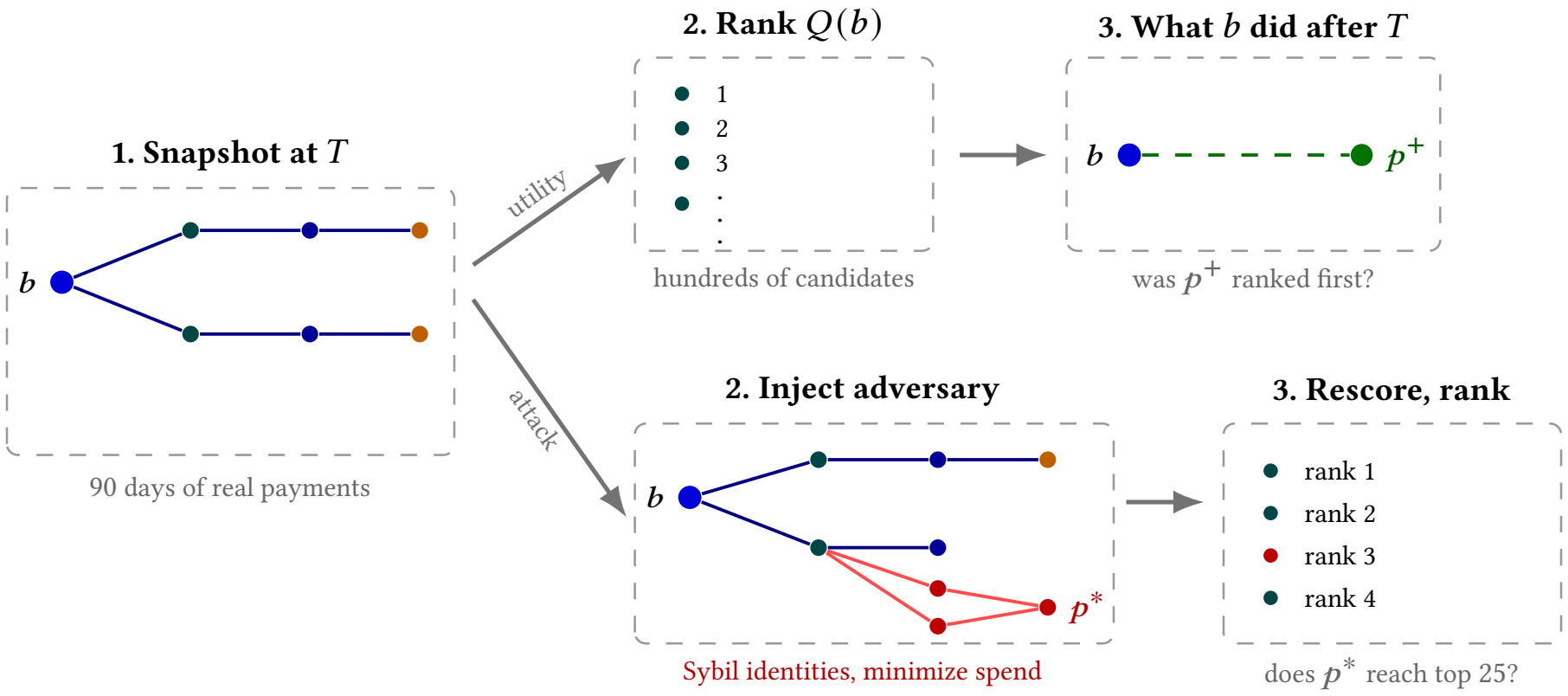 
\subsection{Utility}
\label{sec:p-acc}

We evaluate ranking accuracy using pairwise discrimination ($\mathrm{AUC}$) and top-1 precision ($P@1$), as defined in \Cref{def:utility}; results appear in \Cref{tab:acc}.
$\mathrm{AUC}$ measures the probability of ranking true next provider $p^+$ above an unchosen candidate provider $q \in Q(b) \setminus \{p^+\}$.
$P@1$ measures the fraction of buyers for whom the true choice achieves first rank.

\begin{table}[!tb]
\centering\small
\setlength{\tabcolsep}{4pt}
\begin{tabular}{lrrrrrr}
\toprule
& \multicolumn{2}{c}{$\Rflow$} & \multicolumn{2}{c}{$\Rsum$} & \multicolumn{2}{c}{$\Ramp$} \\
\cmidrule(lr){2-3}\cmidrule(lr){4-5}\cmidrule(lr){6-7}
$w(b,p)$ & AUC & $P@1$ & AUC & $P@1$ & AUC & $P@1$ \\
\midrule
dollars ($w_{\$}$)              & $0.747$ & $4.8$ & $0.859$ & $19.4$ & $0.859$ & $22.5$ \\
dollars $+$ fee ($w_{\$\epsilon}$) & $0.758$ & $4.8$ & $0.871$ & $20.0$ & $0.869$ & $22.8$ \\
$\log$ dollars ($w_{\log}$)     & $0.749$ & $5.2$ & $0.861$ & $20.4$ & $0.862$ & $\mathbf{23.6}$ \\
\textbf{payments} ($w_{\#}$)    & $\mathbf{0.761}$ & $\mathbf{5.4}$ & $\mathbf{0.878}$ & $\mathbf{21.9}$ & $\mathbf{0.877}$ & $23.5$ \\
\midrule
\emph{Baselines} & AUC & $P@1$ & & & & \\
$\Ruse$   & $0.807$ & $3.7$  & \multicolumn{4}{l}{payments into $p'$} \\
$\Rspend$ & $0.749$ & $12.1$ & \multicolumn{4}{l}{co-user spend at $p'$} \\
$\Rcf$    & $0.880$ & $23.2$ & \multicolumn{4}{l}{co-users paying $p'$} \\
\bottomrule
\end{tabular}
\caption{Ranking utility across evaluated buyers under varying edge weights. Naive baselines do not vary by edge weight and are evaluated on the candidate provider set produced by the plain dollar weight.}
\label{tab:acc}
\end{table}

Unlike the flow constructions, the baselines do not depend on edge weight functions---each assigns a score to a provider independently of the querying buyer, so they produce a single ranking {independent of} $w(b,p)$.

Transaction counting ($w_{\#}$) achieves the strongest AUC among flow constructions ($\Rsum\ 0.878$, $\Ramp\ 0.877$, $\Rflow\ 0.761$), approaching the collaborative {filtering} $\Rcf$ baseline ($0.880$).
All constructions exceed the random baseline ($\mathrm{AUC} = 0.5$).
Plain dollars ($w_{\$}$) perform worst because most edges are dominated by a single transaction's value: $51.1\%$ of buyer--provider pairs share only one payment, and on $59.3\%$ of all edges one payment accounts for more than $90\%$ of total edge weight, so the weight reflects that transaction's size rather than relationship strength.

$\Ramp$ achieves the highest top-1 precision---$23.6\%$ under $w_{\log}$---exceeding all baselines.
$\Rflow$ is an outlier: despite $\mathrm{AUC} \ge 0.75$, its $P@1$ never exceeds $5.4\%$, as its minimum cut bottlenecks on the buyer's outgoing arcs and produces multi-candidate ties for $78.0\%$ of buyers.

\paragraph{Impact of Look-Back Window ($\tau$)}
We evaluate look-back $\tau \in [15, 90]$ days on the same set of buyers for $\Ramp$.
Extending $\tau$ monotonically increases $\mathrm{AUC}$ from $0.829$ at $15$ days to $0.869$ at $90$ days.
$P@1$ peaks at $30$ days ($24.7\%$) before candidate provider set expansion ($304 \to 381$) increases rank competition too much.
A 30-to-90 day window balances coverage and top-1 precision (more detail in \Cref{app:tau-sweep}).

\subsection{Attack Cost}
\label{sec:p-attack}

We evaluate economic attack resilience using fee-inclusive dollar weights ($w_{\$\epsilon}$).
We set base fee $\epsilon = \$0.017$, the average per-transaction settlement fee across the chains settling x402 (however, the results are insensitive to the specific value).
Costs are reported in two forms: absolute USD and as a fraction of the total spend of all 813 buyers combined ($\$41{,}331$){, so that attack costs can be read relative to the economic scale of the market being targeted.}

\begin{table}[!tb]
\centering\footnotesize
\setlength{\tabcolsep}{3pt}
\begin{tabular}{lrrrrrrrr}
\toprule
& \multicolumn{2}{c}{top $1$} & \multicolumn{2}{c}{top $5$}
& \multicolumn{2}{c}{top $10$} & \multicolumn{2}{c}{top $25$} \\
\cmidrule(lr){2-3}\cmidrule(lr){4-5}\cmidrule(lr){6-7}\cmidrule(lr){8-9}
& frac. & cost & frac. & cost & frac. & cost & frac. & cost \\
\midrule
\multicolumn{9}{l}{\emph{Baselines}}\\
\quad $\Ruse^{\dagger}$ & $5.263$ & $\$217{,}539$ & $0.178$ & $\$7{,}350$ & $0.021$ & $\$848$ & $0.008$ & $\$310$ \\
\quad $\Rspend$ & $0.000$ & $\$0.02$ & $0.000$ & $\$0.02$ & $0.000$ & $\$0.02$ & $0.000$ & $\$0.02$ \\
\quad $\Rcf$    & $0.009$ & $\$358$ & $0.003$ & $\$138$ & $0.002$ & $\$100$ & $0.001$ & $\$50$ \\
\midrule
\multicolumn{9}{l}{\emph{\sysname Constructions}}\\
\quad $\Rflow$ & $0.113$ & $\$4{,}673$ & $\mathbf{0.059}$ & $\mathbf{\$2{,}420}$ & $\mathbf{0.041}$ & $\mathbf{\$1{,}701}$ & $\mathbf{0.022}$ & $\mathbf{\$898}$ \\
\quad $\Rsum$  & $0.105$ & $\$4{,}330$ & $0.034$ & $\$1{,}420$ & $0.023$ & $\$961$ & $0.012$ & $\$494$ \\
\quad $\Ramp$  & $\mathbf{0.200}$ & $\mathbf{\$8{,}259}$ & $0.040$ & $\$1{,}661$ & $0.028$ & $\$1{,}145$ & $0.013$ & $\$554$ \\
\bottomrule
\end{tabular}
\caption{Minimum cost to capture top-$k$ rank across all buyers, reported in USD and as a fraction of {combined spend of the $813$ tested buyers} ($\$41{,}331$). $^{\dagger}\Ruse$ scores providers globally.}
\label{tab:cover}
\end{table}

Co-user spend ($\Rspend$) costs only a single fee ($\$0.017$) via circular self-payments.
Usage ranking ($\Ruse$) is expensive at rank 1 ($\$217{,}539$) due to requiring $12.8\text{M}$ fees, but falls drastically at subsequent ranks: top-5 ($\$7{,}350$), top-10 ($\$848$), and top-25 ($\$310$).
Collaborative filtering ($\Rcf$) is 23$\times$ cheaper to subvert than $\Ramp$ ($\$358$ vs. $\$8{,}259$ for top-1 market capture).
The flow constructions maintain meaningful costs across all target ranks.
At the loosest threshold examined (top-25), an attacker must still spend between $\$494$ ($\Rsum$) and $\$898$ ($\Rflow$) to reach all buyers, against just $\$50$ under $\Rcf$.
This gap narrows as $k$ increases but does not disappear.

\paragraph{Targeted Manipulation.}
The preceding analysis prices capturing every buyer simultaneously, but an adversary may instead target only the most valuable ones.
Subverting the single largest buyer costs $\$291$ under any construction ($2.3\%$ of that buyer's spend).
Extending to the top decile ($79$ buyers holding $91\%$ of aggregate spend) costs only $\$53$ under $\Rcf$, significantly less than \sysname constructions, which require $\$3{,}183$ ($\Rsum$), $\$4{,}096$ ($\Rflow$), and $\$5{,}490$ ($\Ramp$).
Further detail appears in \Cref{app:targeted-attack} (\Cref{tab:targeted}).

\paragraph{Overall.}

\definecolor{qAnduin}{HTML}{0D9488}
\definecolor{qExisting}{HTML}{EA580C}
\definecolor{qInk}{HTML}{1F2937}
\definecolor{qMuted}{HTML}{6B7280}
\begin{figure}[!tb]
\centering
\resizebox{0.72\linewidth}{!}{%
\begin{tikzpicture}[x=1cm,y=0.6cm,
  pt/.style={circle,fill=qAnduin,inner sep=0pt,minimum size=8pt},
  ptx/.style={rectangle,fill=qExisting,inner sep=0pt,minimum size=7pt},
  lbl/.style={font=\small\bfseries,text=qInk,inner sep=2pt},
  note/.style={font=\scriptsize,text=qMuted,inner sep=1pt},
  qcorner/.style={font=\footnotesize\itshape,text=qMuted,align=center},
  ax/.style={-{Latex[length=2.4mm]},draw=qInk,thick},
]
\fill[qAnduin!12] (5,4) rectangle (10,10);
\draw[qMuted!55,dashed] (5,0) -- (5,10);
\draw[qMuted!55,dashed] (0,4) -- (10,4);
\draw[ax] (0,0) -- (10.6,0);
\draw[ax] (0,0) -- (0,10.6);
\node[font=\small\bfseries,text=qInk] at (5.3,-0.75) {Recommendation utility (accuracy) $\longrightarrow$};
\node[font=\small\bfseries,text=qInk,rotate=90] at (-0.9,5) {Cost to manipulate $\longrightarrow$};
\node[note] at (1.4,-0.4) {low}; \node[note] at (9.7,-0.4) {high};
\node[note,rotate=90] at (-0.4,1.2) {low}; \node[note,rotate=90] at (-0.4,9.5) {high};
\node[qcorner] at (7.5,9.4) {\textbf{Ideal}\\Accurate \& expensive to fake};
\node[qcorner] at (2.2,7.0) {Less accurate\\ but expensive to fake};
\node[qcorner] at (2.2,2.9) {Weak on both};
\node[qcorner] at (7.5,2.9) {Accurate but cheap to fake};
\node[ptx] (u)  at (1.3,9.3) {}; \node[lbl,above=2pt of u] {\Ruse};
\node[note,anchor=west] at (1.75,8.75) {bad accuracy ($3.7\%$)};
\node[pt]  (f)  at (1.7,5.6) {}; \node[lbl,right=2pt of f] {\Rflow};
\node[ptx] (sp) at (4.2,0.8) {}; \node[lbl,above=2pt of sp] {\Rspend};
\node[ptx] (cf) at (8.8,1.2) {}; \node[lbl,above=2pt of cf] {\Rcf};
\node[note] at (8.5,0.5) {\$358 to attack};
\node[pt]  (s)  at (7.3,5.2) {}; \node[lbl,below=2pt of s] {\Rsum};
\node[pt]  (a)  at (8.7,6.7) {}; \node[lbl,above=2pt of a] {\Ramp};
\node[note] at (8.7,6.02) {\$8{,}259 to attack};
\node[pt,minimum size=7pt] (lg1) at (0.4,-2.3) {}; \node[font=\scriptsize,text=qInk,right=2pt of lg1] {\sysname (this paper): \Rflow, \Rsum, \Ramp};
\node[ptx,minimum size=6pt] (lg2) at (6.0,-2.3) {}; \node[font=\scriptsize,text=qInk,right=2pt of lg2] {Existing approaches: \Rcf, \Ruse, \Rspend};
\end{tikzpicture}}
\caption{{Utility versus manipulation cost. Horizontal: ranking accuracy ($P@1$). Vertical: cost to capture rank~1 for all $813$ buyers, as a fraction of their combined spend, under $w_{\$\epsilon}$. Positions are ordinal, not to scale; exact values are in \Cref{tab:acc,tab:cover}. Only $\Rsum$ and $\Ramp$ are both accurate and expensive to subvert. $\Rcf$ matches their accuracy but costs $\$358$ to subvert, and $\Ruse$ is expensive only because it ranks poorly ($P@1 = 3.7\%$).}}
\label{fig:utility-cost}
\end{figure}
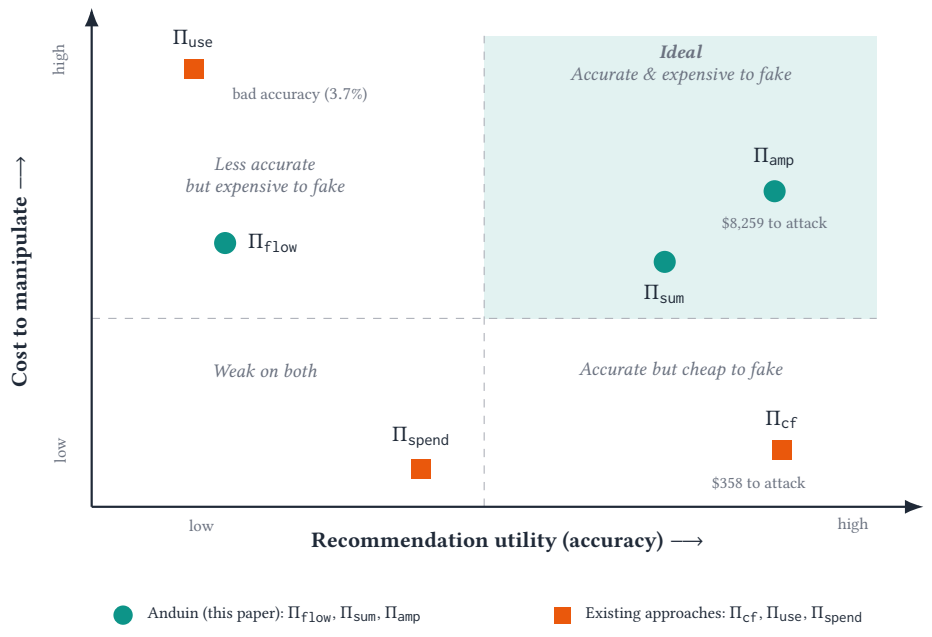

Evaluating utility and attack cost jointly (\Cref{fig:utility-cost}) establishes $\Ramp$ as best construction: baselines either collapse under cheap manipulation ($\Rcf$) or yield poor predictive accuracy ($\Ruse$), whereas $\Ramp$ achieves high precision ($P@1 = {22.8}\%$) while enforcing an attack cost equal to $20\%$ of target buyer spend{, both under $w_{\$\epsilon}$; with $\Rsum$, it is the only scheme in the figure's accurate-and-costly quadrant}.

 \subsection{Extended Empirical Analyses}
\label{app:eval}

\subsubsection{Look-Back ($\tau$) Sensitivity}
\label{app:tau-sweep}

We evaluate the impact of historical look-back duration $\tau \in [15, 90]$ days on candidate provider reachability and predictive precision using the same 813 buyers evaluated in \Cref{sec:methodology}, held fixed across all window sizes so that differences reflect only the rating's look-back, not cohort composition. \Cref{tab:tau-controlled} summarizes performance metrics across window sizes under fee-inclusive dollar weighting ($w_{\$\epsilon}$).

\begin{table}[t]
\centering\small
\setlength{\tabcolsep}{6pt}
\begin{tabular}{@{}r r r r r r@{}}
\toprule
& & & \multicolumn{3}{c}{$\Ramp$} \\
\cmidrule(lr){4-6}
$\tau$ (days) & candidates & coverage & AUC & $P@1$ & $P@1 \mid$ reachable \\
\midrule
$15$ & $261$ & $70.5\%$ & $0.829$ & $12.9$ & $18.3$ \\
$30$ & $304$ & $83.4\%$ & $0.860$ & $\mathbf{24.7}$ & $\mathbf{29.6}$ \\
$45$ & $337$ & $87.2\%$ & $0.860$ & $21.2$ & $24.3$ \\
$60$ & $356$ & $88.4\%$ & $0.863$ & $21.5$ & $24.3$ \\
$75$ & $368$ & $89.7\%$ & $0.868$ & $23.0$ & $25.7$ \\
$90$ & $381$ & $\mathbf{90.7\%}$ & $\mathbf{0.869}$ & $22.8$ & $25.1$ \\
\bottomrule
\end{tabular}
\caption{Look-back window $\tau$ vs. utility on the fixed 813-buyer cohort under fee-inclusive dollar weighting ($w_{\$\epsilon}$). \emph{Candidates}: mean length of candidate list $Q(b)$. \emph{Coverage}: fraction of buyers whose true next provider appears somewhere in $Q(b)$; buyers outside coverage always count as misses. $P@1 \mid \text{reachable}$: top-1 precision conditioned on the true provider being reachable, isolating ranking quality from candidate generation quality.}
\label{tab:tau-controlled}
\end{table}

\emph{Candidates} is the mean number of providers on each buyer's candidate list $Q(b)$; it grows with $\tau$ as more co-users become visible and open new reachable providers.
\emph{Coverage} is the fraction of buyers for whom the true next provider appears anywhere in $Q(b)$; if it is absent, no rating can rank it first.
$P@1 \mid \text{reachable}$ conditions only on buyers whose true provider was reachable, separating ranking quality from candidate generation quality.
Extending $\tau$ monotonically increases both coverage (from $70.5\%$ at 15 days to $90.7\%$ at 90 days) and ranking separation ($\mathrm{AUC}$ reaches $0.869$).
However, $P@1$ peaks at $\tau = 30$ days ($24.7\%$ overall, $29.6\%$ conditional on reachability): while longer windows capture richer history, they also expand candidate provider lists ($304$ at 30 days vs.\ $381$ at 90 days), increasing competition at rank 1.
In practice, a look-back of 30 to 90 days balances coverage against top-1 precision.

\subsubsection{Targeted Attack Cost}
\label{app:targeted-attack}

In addition to systemic market-wide attacks (\Cref{tab:cover}), we evaluate the expenditure required to capture top rank for specific sub-cohorts of high-value buyers.
\Cref{tab:targeted} breaks down the minimum attack cost to achieve rank 1 when targeting the top-$N$ highest-spending buyers (ranked by 90-day transaction volume).

\begin{table}[t]
\centering\small
\setlength{\tabcolsep}{6pt}
\begin{tabular}{@{}r r r r r r@{}}
\toprule
top $N$ & target spend & $\Rcf$ & $\Rflow$ & $\Rsum$ & $\Ramp$ \\
\midrule
$1$   & $\$12{,}612$ & $\$1.86$ & $\$289$     & $\$289$     & $\mathbf{\$291}$ \\
$10$  & $\$22{,}981$ & $\$14$   & $\$2{,}508$ & $\$1{,}716$ & $\mathbf{\$4{,}138}$ \\
$79$  & $\$37{,}566$ & $\$53$   & $\$4{,}096$ & $\$3{,}183$ & $\mathbf{\$5{,}490}$ \\
$813$ & $\$41{,}331$ & $\$358$  & $\$4{,}673$ & $\$4{,}330$ & $\mathbf{\$8{,}259}$ \\
\bottomrule
\end{tabular}
\caption{Targeted attack cost on top-$N$ buyers ranked by spend (rank 1, fee-inclusive dollar weight $w_{\$\epsilon}$, $\epsilon = \$0.017$).}
\label{tab:targeted}
\end{table}

Under collaborative filtering ($\Rcf$), an adversary can capture the single highest-spending buyer for $\$1.86$ and the entire top decile ($N=79$, representing $91\%$ of total market volume) for just $\$53$.
In contrast, flow-based constructions enforce substantial capital requirements.
Subverting the top decile under $\Ramp$ requires $\$5{,}490$ ($14.6\%$ of spend), $\$4{,}096$ under $\Rflow$, and $\$3{,}183$ under $\Rsum$.
This confirms that flow-based ratings effectively price systemic market capture rather than single-buyer preferences.

\subsubsection{Alternative Edge Weighting: Calendar Days ($w_{\mathrm{d}}$)}
\label{app:returndays}

An alternative edge weight deferred from \Cref{sec:weights} counts distinct calendar days with settled transactions: $w_{\mathrm{d}}(b,p) = |\{\mathrm{day}(t_i)\}|$. In machine-to-machine commerce where an automated agent can execute hundreds of micro-transactions within seconds, counting distinct return days filters single-day payment bursts and captures sustained counterparty relationships.

Empirically, $w_{\mathrm{d}}$ yields the highest ranking utility across all evaluated metrics, lifting $\Ramp$ to $\mathrm{AUC} = 0.881$ and $P@1 = 25.5\%$ (compared to $0.869$ and $22.8\%$ under $w_{\$\epsilon}$), and lifting $\Rsum$ by a comparable margin. 

However, we omit $w_{\mathrm{d}}$ from our primary attack cost benchmarking because temporal delay cannot be directly priced in financial terms. Forging a unit of $w_{\mathrm{d}}$ requires burning a transaction fee \emph{and waiting a calendar day}. While time imposes a genuine physical constraint on adversaries, expressing attack costs as a fraction of calendar days creates an incommensurable comparison against dollar-denominated bounds.
 \section{Discussion}
\label{app:discussion}

This section details considerations, limitations, and design extensions for graph-based agentic reputation. 

\paragraph{Cold-Start Dynamics}
\label{app:cold-start}
A buyer without prior transaction history has no known providers $K(b)$, co-users $C(b)$, or reachable candidate providers $Q(b)$. In this initial state, graph ratings are undefined across all providers, much like in collaborative filtering~\cite{schein2002methods}. In agentic marketplaces, new buyers may initially bootstrap via curated allow-lists~\cite{coinbase2026bazaar,circle2026agentstack}. Our graph framework improves naturally as transaction history accumulates: even a single settled payment populates $K(b)$ and establishes a personalized recommendation. Similarly, newly listed providers remain unrated until adopted by at least one co-user.

\paragraph{Multi-Agent Wallets}
\label{app:multi-agent}
When a single onchain address acts as a shared treasury for multiple distinct AI agents or providers, their transaction histories are conflated. This pools unrelated co-users and buyers, introducing noise to our recommendation. Adopting per-agent sub-account keys or signing metadata (e.g., ERC-4337 session keys or BIP-32 wallet key derivation conventions) would allow our payment graph to decompose multi-agent wallets into distinct vertices, aligning our graph representation with underlying agent intent.

 \section{Conclusion}
\label{sec:conclusion}
This paper demonstrates that historically settled onchain payments can be harnessed to rank service providers in autonomous agentic markets without protocol modifications, external credentials, or manual curation. Using a 157-day continuous settlement dataset of $66.1\text{M}$ payments from the x402 marketplace, we modeled transactions as a bipartite payment graph and formalized the dual criteria of predictive accuracy (\emph{utility}) and bounded expenditure (\emph{attack cost}).

\sysname comprises three rating constructions derived from this graph model: $\Rflow$ which enforces strict rating ceilings on the buyer's historical spend, $\Rsum$ which replicates capacity per co-user to improve rating efficacy and $\Ramp$ which pools capacity across co-users aligned on candidate transactions. Together, they demonstrate that localized graph flow can replace ad-hoc reputation heuristics while enforcing a market-scaled economic cost on Sybil manipulation.

\bibliographystyle{ACM-Reference-Format}

\appendix
\renewcommand{\theHsection}{appendix.\Alph{section}}%
\section{Identification of x402 Settlements}
\label{app:x402-identification}

We identify x402 settlements on EVM networks and Solana using a combination of event logs,
transaction structure, and known facilitator and seller addresses.

\subsection{EVM Settlements}

We use the following rules to identify x402 transactions involving ERC-20
transfers of USDC and USDT:

\begin{center}
\begingroup
\footnotesize
\setlength{\tabcolsep}{3pt}
\renewcommand{\arraystretch}{1.15}
\begin{tabular}{@{}p{0.20\linewidth} p{0.29\linewidth} p{0.45\linewidth}@{}}
\hline
{\raggedright\textbf{Settlement method}\par} &
{\raggedright\textbf{Event logs}\par} &
{\raggedright\textbf{Top-level transaction \texttt{to} address}\par} \\
\hline
{\raggedright EIP-3009 direct\par} &
{\raggedright\texttt{AuthorizationUsed} and \texttt{Transfer}\par} &
{\raggedright Token contract.\par} \\
{\raggedright Permit2 Exact\par} &
{\raggedright\texttt{Settled} or \texttt{SettledWithPermit}, and
\texttt{Transfer}\par} &
{\raggedright Permit2 Exact proxy:
{\scriptsize\texttt{\seqsplit{0x402085c248eea27d92e8b30b2c58ed07f9e20001}}}.\par} \\
{\raggedright Permit2 Upto\par} &
{\raggedright\texttt{Settled} or \texttt{SettledWithPermit}, and
\texttt{Transfer}\par} &
{\raggedright Permit2 Upto proxy:
{\scriptsize\texttt{\seqsplit{0x4020a4f3b7b90cca423b9fabcc0ce57c6c240002}}}.\par} \\
{\raggedright Multicall\par} &
{\raggedright A settlement event listed above and \texttt{Transfer}\par} &
{\raggedright Neither the token contract nor either Permit2 proxy; a residual
category for routers, batchers, and other intermediaries.\par} \\
\hline
\end{tabular}
\endgroup
\end{center}

Permit2 Exact and Upto emit the same settlement events and are distinguished
by their x402-specific proxy addresses. These addresses provide sufficient
evidence of x402 settlement without further facilitator or seller validation.
By contrast, \texttt{AuthorizationUsed} is a generic EIP-3009 event. Direct
EIP-3009 and Multicall candidates are therefore retained only when the
top-level transaction \texttt{from} address is a known facilitator or the
transfer recipient is a known x402 seller. Each qualifying \texttt{Transfer}
log produces one record containing the buyer, seller, and amount, with the
top-level transaction \texttt{from} address identifying the facilitator.

\subsection{Solana Settlements}

On Solana, we identify outgoing stablecoin transfers in which the fee payer
differs from the transfer's source authority. A transfer is retained when
either the fee payer is a known facilitator or the destination is a known
x402 seller.

\subsection{Facilitator and Seller Sources}

The initial list of known facilitator addresses is drawn from two public Dune
dashboards and their underlying queries: x402 Analytics\footnote{\url{https://dune.com/hashed_official/x402-analytics}},
published by Hashed, and x402 Facilitators\footnote{\url{https://dune.com/dune/x402-facilitators}}, published
by Dune. On EVM networks, the list is supplemented with addresses observed submitting
settlements through the x402 Permit2 proxies. Known sellers are sourced from
Coinbase's public x402 discovery registry\footnote{\url{https://api.cdp.coinbase.com/platform/v2/x402/discovery/resources}}, which publishes the payment addresses
accepted by registered resources on each network.

\subsection{Limitations}

EIP-3009, Multicall, and Solana transactions involving both an unknown facilitator and an unregistered seller are not captured.
 
\end{document}